%% file: main.tex
\documentclass[10pt,twocolumn,letterpaper]{article}
\usepackage[preprint]{cvpr} 
\usepackage{float}

\definecolor{cvprblue}{rgb}{0.21,0.49,0.74}
\usepackage[pagebackref,breaklinks,colorlinks,allcolors=cvprblue]{hyperref}
\usepackage{multirow}
\usepackage{makecell} 
\usepackage{animate}
\usepackage{graphicx}
\usepackage{xcolor}
\usepackage[table]{xcolor}
\usepackage[misc]{ifsym}
\usepackage{footnote}
\usepackage{pifont}
\usepackage{needspace}
\usepackage{caption}
\definecolor{lavender}{RGB}{230,230,250}
\definecolor{lightgreen}{RGB}{220, 255, 220}
\newcommand{\cmark}{\ding{51}}
\newcommand{\xmark}{\ding{55}}
\usepackage{amsmath}
\usepackage{booktabs}
\usepackage{subcaption}
\usepackage{wrapfig}
\usepackage{comment}
\newcommand{\model}{LinkVLA\xspace}
\newcommand{\modelit}{\textit{LinkVLA}\xspace}
\newcommand{\modelbf}{\textit{\textbf{LinkVLA}}\xspace}
\usepackage{array}
\usepackage{tabularx}
\newcolumntype{Y}{>{\centering\arraybackslash}X}
\newcolumntype{C}[1]{>{\centering\arraybackslash}p{#1}}
\newcolumntype{L}[1]{>{\raggedright\arraybackslash}p{#1}}
\newcolumntype{G}[1]{>{\columncolor{gray!17}\centering\arraybackslash}p{#1}}

\title{Unifying Language-Action  Understanding and Generation \\ for Autonomous Driving}

\author{
    Xinyang Wang$^{1,2, *}$\quad
    Qian Liu$^{2, *}$\quad
    Wenjie Ding$^{2, *}$\quad
    Zhao Yang$^{2, \dagger}$\quad
    Wei Li$^{2}$ \\
    Chang Liu$^{2}$\quad
    Bailin Li$^{2}$\quad
    Kun Zhan$^{2}$\quad
    Xianpeng Lang$^{2}$\quad
    Wei Chen$^{1}$ \\[2mm]
    $^1$State Key Lab of CAD\&CG, Zhejiang University \quad $^2$Li Auto
}
\usepackage{xcolor}
\begin{document}
\maketitle

{
  \renewcommand{\thefootnote}{}
  \footnotetext{Work done while interning at Li Auto.}
  \footnotetext{$^*$ Equal contribution. $^\dagger$ Project Lead.}
}
\input{sec/0_abstract.tex}
\input{sec/1_intro.tex}
\input{sec/2_related.tex}
\input{sec/3_method.tex}
\input{sec/4_exp.tex}

\input{sec/5_conclusion.tex}
\input{sec/6_ref.tex}
\input{sec/7_appendix.tex}

\end{document}

%% file: sec/0_abstract.tex
\begin{abstract}
Vision-Language-Action (VLA) models are emerging as a promising paradigm for end-to-end autonomous driving, valued for their potential to leverage world knowledge and reason about complex driving scenes. 
However, existing methods suffer from two critical limitations: a persistent misalignment between language instructions and action outputs, and the inherent inefficiency of typical auto-regressive action generation.
In this paper, we introduce \modelbf, a novel architecture that directly addresses these challenges to enhance both alignment and efficiency.
First, we establish a structural link by unifying language and action tokens into a shared discrete codebook, processed within a single multi-modal model. This structurally enforces cross-modal consistency from the ground up. 
Second, to create a deep semantic link, we introduce an auxiliary action understanding objective that trains the model to generate descriptive captions from trajectories, fostering a bidirectional language–action mapping.
Finally, we replace the slow, step-by-step generation with a two-step coarse-to-fine generation method~($C2F$) that efficiently decodes the action sequence, saving $86\%$ inference time.
Experiments on closed-loop driving benchmarks show consistent gains in instruction following accuracy and driving performance, alongside reduced inference latency.
\end{abstract}

%% file: sec/1_intro.tex
\section{Introduction}
\label{sec:intro}
\begin{figure}[t]\centering
\subfloat[Time \textit{vs.} performance.\label{fig:intro_1}]
{\includegraphics[width=0.9\linewidth]{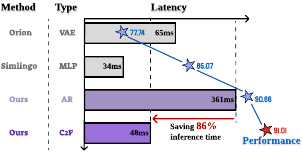}}\hfill
\vspace{5pt}
\subfloat[Visualization of instruction-following capability.\label{fig:intro_2}]
{\includegraphics[width=0.9\linewidth]{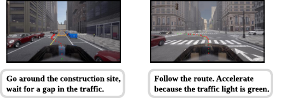}}
\vspace{-5pt}
\caption{\model achieves both higher performance and lower latency in closed-loop evaluation and equipped with superior instruction-following capability.}
\vspace{-8pt}
\label{fig:intro}
\end{figure}
End-to-end learning~\cite{hu2023planning,jiang2023vad,chen2024vadv2,sun2025sparsedrive,liao2025diffusiondrive,xing2025goalflow,weng2024drive,chitta2022transfuser, li2024hydra, hu2022st, zheng2024occworld, zheng2024genad} emerged as a dominant paradigm for developing autonomous driving systems, learning direct sensorimotor policies from raw sensor inputs to vehicle planning. While conventional end-to-end models excel at reactive control in familiar scenarios, they often struggle with complex reasoning, long-tail events, and human interaction~\cite{sima2024drivelm, tian2024drivevlm}.
Vision-language-models~(VLMs)~\cite{jaech2024openai, guo2025deepseek} have attracted significant attention in recent years due to their  capable of leveraging extensive world knowledge and sophisticated reasoning capabilities and have been introduced into driving scenarios to enhance their generalization~\cite{wang2023drivemlm, chen2025automated, li2024driving, wang2025omnidrive, ma2024dolphins, winter2025bevdriver, tian2025nuscenes, mao2023gpt, jin2023adapt, park2024vlaad, ding2024hint, marcu2024lingoqa, jiang2025alphadrive, jiang2024senna}. 
Vision-Language-Action~(VLA) methods extend VLMs to action genaration and have been explored in robotics~\cite{pertsch2025fast, black2410pi0} and autonomous driving~\cite{zhou2025autovla,li2025recogdrive,fu2025orion,renz2025simlingo,li2025drivevla,zeng2025futuresightdrive, jiang2025diffvla, xu2024drivegpt4, shao2024lmdrive, zhou2025opendrivevla, waywe2024lingo} to generate physical feasible action  based on visual and language inputs. VLAs promise a paradigm shift from learning implicit, reactive policies to developing agents capable of explicit reasoning and long-horizon planning. These models unlock superior interactivity and flexible instruction following, allowing language to serve as a powerful medium for conveying explicit rules and compositional logic, representing a critical step towards achieving more generalist and trustworthy autonomous agents.

At the heart of the VLA paradigm lies the ability to follow natural language instructions. Instruction following is a critical function for real-world deployment, enabling dynamic re-tasking and enhanced user trust through transparent interaction. However, a fundamental challenge plagues current VLA models: a persistent misalignment between language understanding and physical action~\cite{renz2025simlingo, glossop2025cast, wang2025omnidrive}. A model might correctly generate the decision \textit{change lane to the left}, yet output a lane keep trajectory. This failure in instruction following undermines the core promise of VLAs and poses a significant barrier to their safety and reliability.

Several lines of research have emerged to address the misalignment problem. Some approaches focus on improving data collection techniques~\cite{renz2025simlingo, glossop2025cast, wang2025omnidrive}, a strategy that sidesteps the fundamental modeling challenge. Others rely on post-hoc policy refinement using reinforcement learning~\cite{zhou2025autovla}, which treats alignment as a corrective afterthought. A third vein attempts to align modalities through implicit distribution matching in latent space~\cite{fu2025orion}, which lacks a direct, verifiable supervision.
In contrast to these methods, we argue that this semantic gap necessitates an explicit bidirectional link woven into the primary supervised learning phase. To this end, we introduce \modelbf, a novel VLA model designed specifically to strengthen this connection. 

First, we establish a structural link at the architectural level. By unifying language instructions and trajectories into a shared discrete codebook, we eliminate the modality gap by design, forcing both concepts into a common representational ground. Second, it forges a deep semantic link through a novel bidirectional learning objective. We introduce an explicit action-understanding objective that compels the model to translate the planned actions back into descriptive text. This synergy enforces bidirectional consistency, ensuring the link between language and action is formative and verifiable. While this tightly coupled representation is powerful for alignment, its auto-regressive nature creates an inference bottleneck. To make our framework practical, we replace conventional step-by-step decoding with a coarse-to-fine, two-step generation process. This mechanism first generates a high-level structural outline of the trajectory and then refines this outline into the full, fine-grained path. As visualized in Figure~\ref{fig:intro}, \modelit not only achieves a dramatic reduction in inference latency but also achieves consistent gains in instruction following accuracy and driving performance. Here are the main contributions:

\begin{itemize}
\item \textit{A unified tokenized framework} that learns a shared codebook for language and action, structurally bridging the modality gap to enhance alignment.
\item \textit{An explicit action understanding objective} that enforces bidirectional semantic consistency.
\item \textit{A coarse-to-fine action generation schema} that drastically reduces inference latency.
\item \textit{State-of-the-art performance} on challenging closed-loop driving benchmarks, demonstrating significant gains in both instruction-following accuracy and driving ability.
\end{itemize}

%% file: sec/2_related.tex
\section{Related Work}\label{sec:related}

\subsection{End-to-end Autonomous Driving}
End-to-end autonomous driving has achieved tremendous success in recent years.
UniAD~\cite{hu2023planning} adopts an ultimately planning-oriented design philosophy that integrates perception, prediction, and planning into a single network.
VAD~\cite{jiang2023vad} models driving scenarios into fully vectorized representations, thereby eliminate computationally intensive rasterized representations and resulting in a much faster running speed.
ParaDrive~\cite{weng2024drive} proposes a fully parallel end-to-end architecture and conducts a comprehensive exploration of the design space of modular stacks.
VADv2~\cite{chen2024vadv2} discretizes the planning action space into a large planning vocabulary and leverages extensive driving demonstrations to learn the probability distribution of planning actions.
SparseDrive~\cite{sun2025sparsedrive} unifies multiple tasks through sparse instance representation and propose a symmetric
sparse perception module and a parallel motion planner.
GenAD~\cite{zheng2024genad} casts autonomous driving into a generative modeling problem.
DriveTransformer~\cite{jia2025drivetransformer} adopts task parallelism and sparse representation to improve efficiency by working in a streaming manner. 
DiffusionDrive~\cite{liao2025diffusiondrive} uses anchor to truncate the diffusion process, achieving better performance, higher inference efficiency.
GoalFlow~\cite{xing2025goalflow} constrains the generated trajectories by introducing a goal point and use it as the condition of flow matching to generate multimodal trajectories.
While proficient at reactive control within known operational domains, these methods display critical limitations in complex logical reasoning, long-tail event processing, and human-interactive task execution.
\vspace{0.4cm}
\begin{figure*}[ht!]
    \centering
    \includegraphics[width=\textwidth]{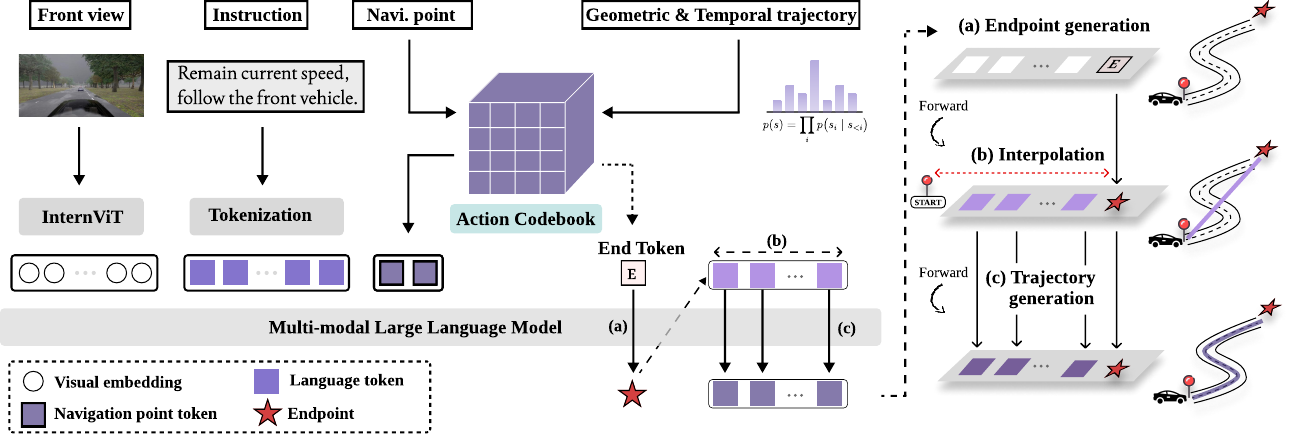}
    \caption{\textbf{An overview of the \modelbf architecture.} The model comprises a pretrained InternViT~\cite{chen2024internvl} visual backbone and a Qwen2-0.5B~\cite{team2024qwen2} LLM. At its core, \modelit unifies language tokens and action tokens (for navigation points and trajectories) into a single, shared codebook. Training is driven by a unified objective for both language-action understanding and generation, ensuring deep semantic alignment. Inference employs an efficient coarse-to-fine process: first, (a) the model predicts the trajectory endpoint, which is then (b) interpolated into coarse waypoints before being (c) refined into the final, smooth trajectory.}
    \label{fig:method}
\end{figure*}
\vspace{0.4cm}
\subsection{VLM and VLA for Autonomous Driving}
Early-stage VLMs primarily follow a language-centric paradigm that use visual question answering to discribe the action~\cite{jin2023adapt, park2024vlaad, marcu2024lingoqa}.
DriveGPT4~\cite{xu2024drivegpt4} employs Large Language Models (LLMs) to generate scene explanations and control signals.
DriveVLM~\cite{tian2024drivevlm} uses VLM to enhance scene understanding and planning capabilities.
Drivemlm~\cite{wang2023drivemlm} standardizes decision states to bridge the gap between the language decisions and the vehicle control signals and uses LLM to make driving decisions and explanations.
EMMA~\cite{hwang2024emma} capitalizes on Gemini’s multimodal capabilities by representing all non-sensor data – including navigation instructions, vehicle status, trajectories, and 3D positions – as textual sequences, thereby transferring pre-trained LLMs’ world knowledge to autonomous driving tasks.

VLA direct outputs action from raw input.
Opendrivevla~\cite{zhou2025opendrivevla} projects visual tokens into a unified semantic space to bridge the modality gap.
DriveMoE~\cite{yang2025drivemoe} introduces vision MoE and action MoE to handle diverse scenarios.
Recogdrive~\cite{li2025recogdrive} injects the VLM’s learned driving priors into a diffusion planner and uses reinforcement learning to  bridge the gap between language and action.
DiffVLA~\cite{jiang2025diffvla} proposes a hybrid sparse-dense diffusion policy empowered by a Vision-Language Model~(VLM), enabling efficient multimodal driving behavior generation.
However, a modality gap still exists, resulting in misalignment between language and action.

\subsection{Language and Action Alignment}
AutoVLA~\cite{zhou2025autovla} unifies reasoning with action generation and employs GRPO~\cite{shao2024deepseekmath} to improve planning performance.
The step-by-step autoregressive action generation in AutoVLA makes it inefficienct in practical deployment.
ORION~\cite{fu2025orion} bridges reasoning and action spaces by merging generative planners with VLM architectures, jointly optimizing VQA and planning.
SimLingo~\cite{renz2025simlingo} focuses on aligning linguistic comprehension with driving actions.
However, a persistent misalignment between language instructions and action still exist in  ORION and SimLingo due to the modal gap.
CAST~\cite{glossop2025cast} leverage vision language models to create counterfactual labels to augment existing robot datasets.
OmniDrive~\cite{wang2025omnidrive} proposes a holistic vision-language dataset for aligning agent models with 3D driving tasks using counterfactual reasoning.
In unified multimodal understanding and generation model, understanding and generation are found to be mutually beneficial to each other~\cite{tong2025metamorph, xie2024show, zhou2024transfusion}.
Inspired by this, we propose a novel architecture that improves action generation ability by introducing an action understanding objective without the need for additional data curation.

%% file: sec/3_method.tex
\section{Method}\label{sec:method}

Our proposed model, \modelbf, is a Vision-Language-Action model designed to enhance language-action alignment and inference efficiency in autonomous driving.
Our methodology introduces three key innovations, as illustrated in Figure~\ref{fig:method}. 
First, we establish a unified auto-regressive framework that models language and action tokens within a single discrete space~(Sec.~\ref{ssec:unified_ar}). 
Second, to enhance semantic alignment, we introduce a novel action understanding objective that fosters a bidirectional mapping between language and trajectories (Sec.~\ref{ssec:action_understanding}). 
Third, we replace slow, sequential decoding with an efficient, coarse-to-fine generation mechanism that drastically reduces inference latency (Sec.~\ref{ssec:hierarchical_generation}).

\subsection{Unified Tokenization Framework}
\label{ssec:unified_ar}

We posit that the language-action misalignment in autonomous driving is a direct consequence of an architectural schism between modalities. To eliminate this, our method is founded on the principle of unification: modeling the entire process—from understanding an instruction to generating a trajectory—within a single unified framework. Our approach maps both the language instruction $L$ and the action trajectory $\mathcal{T}$ into a unified sequence of discrete tokens, which is then processed by a VLM backbone. For language, we leverage the VLM's existing tokenizer. For actions, which are inherently continuous, we devise a {spatial-aware action tokenization} scheme. Instead of regressing continuous values, our model predicts a sequence of action tokens from a tokenized codebook.

\paragraph{Unified Token Space.} Our approach is founded on a unified token space for language and action. We achieve this by first quantizing continuous trajectories: the local Bird's-Eye-View (BEV) space is partitioned into a grid of $K_{action}$ cells, each defining a unique action token. A trajectory $\mathcal{T} = \{w_1, \dots, w_T\}$ is thus converted into a sequence of action tokens ${A} = \{a_1, \dots, a_T\}$ by mapping each waypoint to its corresponding cell. This action codebook ($\mathcal{C}_{action}$) is then merged with the model's text vocabulary (size $K_{text}$) to form a single, unified codebook $\mathcal{C}$ of size $K = K_{text} + K_{action}$. The action token embeddings are learned end-to-end, forcing the model to map linguistic and spatial concepts into a shared representation and enabling a single VLM model to process both. During inference, each predicted action token is simply decoded back to the center of its corresponding grid cell.

\paragraph{Action Tokenization.} 
\label{sec:tokenization}
Naive tokenization, which tokenizes waypoints onto a uniform BEV grid using one-hot labels, suffers from two key issues. First, a uniform grid allocates resolution evenly, failing to provide the fine-grained precision essential for near-field control. Second, the hard assignment of one-hot labels discards the grid's inherent spatial topology, complicating the learning of spatial priors. To mitigate these issues, we introduce two key refinements: Log Coordinate Transformation and Spatial Soft-labeling.

\noindent{1) \textit{Log Coordinate Transformation.} We devise a non-uniform quantization scheme that prioritizes precision near the ego-vehicle. This is achieved by first applying a non-linear transformation to the waypoint coordinates~$(x, y)$ independently along each axis. Specifically, each coordinate $z \in \{x, y\}$ is transformed using a signed logarithmic function:
\begin{equation}
z' = \operatorname{sign}(z) \cdot \log(1 + k \cdot |z|).  
\label{eq:loggrid}
\end{equation} 

Here, $k$ is a positive scaling factor that controls the size of the linear region around the origin. The resulting transformed space $(x', y')$ is then uniformly quantized to produce the action tokens.
\vspace{0.5cm}

\noindent{2) \textit{Spatial Soft-labeling.}
To embed a physical prior about the continuity of the action space into our learning objective, we employ a spatial soft-labeling strategy. A standard one-hot target provides a discrete supervisory signal, but does not explicitly account for the spatial topology of our action grid. Our approach refines this by defining a smooth target distribution that acknowledges spatial adjacency.

Specifically, for a ground-truth token $a_{gt}$, we construct the target distribution $q(a)$ over all action tokens $a \in \mathcal{C}_{action}$ as a normalized 2D Gaussian centered on the coordinates of $a_{gt}$ with radius $R$:
\begin{equation}
    q(a) = \frac{1}{Z} \exp\left(-\frac{\| \text{pos}(a) - \text{pos}(a_{gt}) \|_2^2}{2\sigma^2}\right),
    \label{eq:soft_target}
\end{equation}
where $\text{pos}(a)$ maps an action token to its 2D coordinates in the spatial grid, $\| \cdot \|_2^2$ is the squared Euclidean distance, $\sigma$ is a hyperparameter controlling the spread of the distribution, and $Z$ is a normalization constant ensuring $\sum_{a \in \mathcal{C}_{action}} q(a) = 1$.
For action generation, the model's predicted distribution $p(a)$ is then optimized to match this soft target using the cross-entropy loss:
\vspace{-0.2cm}
\begin{equation}
    \mathcal{L}_{\text{generation}} = - \sum_{a \in \mathcal{C}_{action}} q(a) \log p(a).
    \label{eq:soft_loss}
\end{equation}

This objective encourages the model to assign probability mass not only to the correct token but also to its spatial neighbors, as dictated by the Gaussian shape. This fosters a locally smooth action manifold, making the model more robust to minor groundtruth errors. When incorporating chain-of-thought, an additional standard cross-entropy loss for language generation is added to this objective. 

\subsection{Unified Language-Action Understanding and Generation}
\label{ssec:action_understanding}

To strengthen the link between language and behavior, we introduce a bidirectional training objective inspired by the duality of image captioning and text-to-image generation. In generative vision-language modeling, the tasks of generating text $L$ from an image $V$ ($p(L|V)$) and an image from text ($p({V}|{L})$) are reciprocal~\cite{xie2024show, zhou2024transfusion, tong2025metamorph}. Training a model on both objectives has been shown to produce a more robust and aligned joint embedding space.

\begin{figure}[h]
    \centering
    \includegraphics[width=0.48\textwidth]{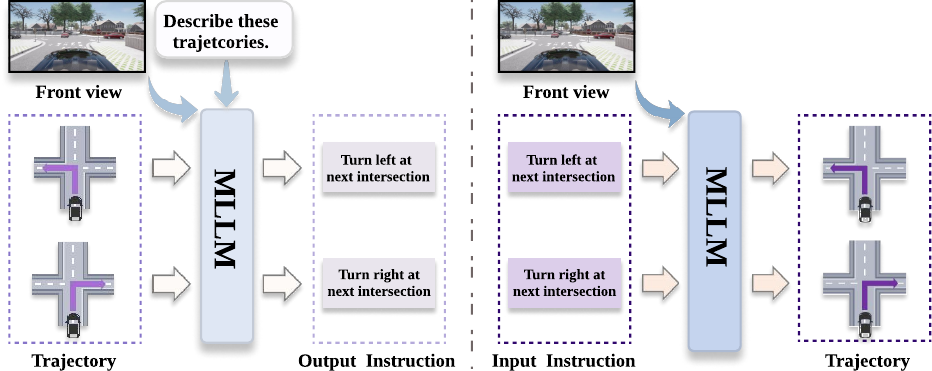}
    \caption{Illustration of the action understanding~(\textbf{Left}) and the action generation~(\textbf{Right}).}
    \label{fig:Align}
\end{figure}

We posit that a similar duality exists for the language-action mapping within VLA models. The conventional task is \textit{action generation}, where a language instruction guides the prediction of an action sequence $A$~($p(A|L)$). This is analogous to text-to-image synthesis. We propose its reciprocal: \textit{action understanding}, where an executed action sequence $A$ is used to infer the original language instruction~($p(L|A)$). This mirrors image captioning, as the model must produce a linguistic description explaining the observed behavior. Crucially, both of these mappings are grounded in a shared visual context $V$, which provides the necessary environmental awareness. Formally, besides $\mathcal{L}_{generation}$~(Eq.~\ref{eq:soft_loss}), we introduce a reciprocal objective designed for reconstructing the language instruction ${L}$ given the vision and action inputs:
\begin{equation}
    \mathcal{L}_{\text{understanding}} = - \sum_{j} \log p(l_j | V, A, l_{<j}),
    \label{eq:actionunderstanding}
\end{equation}
where $l_{<j}$ represents the preceding ground-truth tokens.

The final training objective is a weighted sum of these two losses:
$\mathcal{L}_{\text{total}} = \mathcal{L}_{\text{generation}} + \lambda \mathcal{L}_{\text{understanding}}$, where $\lambda$ is a balancing hyperparameter. By forcing the model to solve this inverse problem, we enforce a bidirectional consistency within the shared embedding space. This process enriches the semantic grounding of the action tokens, ensuring they are intrinsically linked to descriptive linguistic concepts, thereby leading to better instruction following. In practice, both tasks are handled by the same decoder by simply swapping the roles of $L$ and $A$ as the prediction target.

\subsection{Coarse-to-Fine Action Generation}
\label{ssec:hierarchical_generation}

Auto-regressive generation of a long trajectory of $T$ waypoints requires $T$ sequential forward passes, which is computationally expensive and introduces significant inference latency. To address this, we collapse the $T$-step sequential dependency into a two-stage process: (1) endpoint prediction and coarse trajectory initialization, and (2) parallel trajectory refinement. 
Following~\cite{renz2025simlingo}, our action outputs include both temporal speed waypoints and geometric path waypoints. As both are treated as point sequences and processed symmetrically throughout our framework, we will hereafter refer to them collectively as a \textit{trajectory} and its constituents as \textit{waypoints} for clarity. 

\paragraph{Training with a Coarse Trajectory Prior.}
Our coarse-to-fine inference is enabled by a carefully designed training objective. To directly predict endpoints, we put special tokens at the beginning of the decoder's input sequence. 
During training, the ground-truth target sequence is reordered, $\{w_T, w_1, w_2, ..., w_{T-1}\}$, teaching the model to associate the special token with the endpoint prediction.
For the refinement stage, we simulate coarse trajectory via linear interpolation using the groundtruth endpoint, quantize it into coarse waypoint tokens as model inputs, and train the model to map these coarse tokens to the corresponding fine-grained trajectory.

\paragraph{Coarse Trajectory Initialization.} During inference, we first establish a strong structural prior for the trajectory, which serves to guide the subsequent generation steps. With our modified training sequence, the model performs a single forward pass to predict specifically the final waypoint, $\hat{w}_T$. While this initial step is inspired by goal-point prediction methods~\cite{xing2025goalflow}, our approach is fundamentally distinct as we integrate endpoint prediction and trajectory refinement within a single, unified transformer architecture.

Given the start point $w_0$ (the ego-vehicle's origin at (0, 0)) and the predicted end point $w_T$, we construct the coarse trajectory, $\mathcal{W}_{coarse}$, via linear interpolation:
\begin{equation}
    w_i^{coarse} = w_0 + \frac{i}{T}(w_T - w_0) \quad \text{for } i \in \{1, \dots, T\}
    \label{eq:interpolation}
\end{equation}
Then these waypoints are tokenized into trajectory tokens, as an initial scaffold for the subsequent refinement stage.

\paragraph{Parallel Trajectory Refinement.} The second inference step refines the coarse, straight-line path into a dynamically feasible trajectory. 
We formulate this as a structure-preserving refinement, where each coarse waypoint $w_i^{coarse}$ is mapped to its corresponding refined waypoint $w^{fine}_i$. 
Given tokenized coarse waypoints as initial input, \model predicts $T$ refined points in parallel.
Conditioned by the vision-language context via cross-attention, the refined path respects lane boundaries, avoids obstacles, and adheres to the language instruction.

%% file: sec/4_exp.tex
\section{Experiments}\label{sec:exp}
\input{tables/tab_main_ability}

We first introduce the benchmarks, evaluation metrics (Sec.~\ref{e-sec:benchmark}), and implementation details (Sec.~\ref{e-sec:implement}). Then we present and analyze the closed-loop results and the instruction-following ability (Sec.~\ref{e-sec:results}), followed by detailed ablation experiment and comparisons (Sec.~\ref{sec:comparison}).

\subsection{Settings}
\label{e-sec:benchmark}
\noindent\textbf{Bench2Drive.}
We train and evaluate \model using the Bench2Drive benchmark~\cite{jia2024bench}, which provides a set of interactive scenarios within the widely used CARLA simulator~\cite{dosovitskiy2017carla}. 
We follow SimLingo~\cite{renz2025simlingo} and use an open-source expert PDM-lite~\cite{sima2024drivelm} to collect driving dataset in the CARLA simulator.
Our evaluation follows the CARLA v2 closed-loop protocol for end-to-end autonomous driving, comprising 44 interactive scenarios with 5 routes each, for a total of 220 official routes across diverse weather conditions. We report performance using the benchmark's official metrics: Driving Score (DS), Success Rate (SR), Efficiency, Comfortness, and Multi-Ability.

\noindent\textbf{Instruction-following Evaluation.} 
We evaluate the model's instruction-following capabilities using the \textit{Action Dreaming} dataset from SimLingo~\cite{renz2025simlingo}. This dataset is designed to assess a model's ability not only to comprehend scene-specific knowledge from language but also to translate this understanding into the corresponding action space. Given a natural language instruction, the model is expected to generate a sequence of actions that corresponds to the command. The evaluation is conducted on the CARLA Town 13 dreamer dataset for validation. The instructions belong to one of six classes: Slow down, Speed up, Reach target speed, Lane
change, Object centric. 
Performance is measured by the success rate.

\noindent\textbf{DriveLM-hard (VQA) and Commentary.}
We evaluate VQA and commentary generation on the DriveLM-hard benchmark~\cite{renz2025simlingo}. This challenging validation set is derived from DriveLM~\cite{sima2024drivelm} and focuses on the CARLA Town 13 environment. To ensure a balanced test that includes rare cases, the benchmark was constructed by uniformly sampling 10 examples per answer type, rather than using simple random sampling. The final dataset contains 330 VQA answer types and 190 Commentary types. We report scores using the \textit{SPICE}, \textit{BLEU}, and \textit{ROUGE-L} metrics.

\input{tables/tab_latency}
\subsection{Implementation details}
\label{e-sec:implement}
\noindent\textbf{Action Tokenize.}
The framework operates within a Bird's-Eye-View (BEV) space spanning the coordinate ranges $x \in [0, 50]\text{m}$ and $y \in [-30, 30]\text{m}$. To create a discrete action space, these coordinates are first transformed using the logarithmic function detailed in Sec~\ref{sec:tokenization} (with hyperparameter $k=5$) and subsequently discretized into a uniform grid with a 0.1 step size. This process yields a $56 \times 101$ grid, which constitutes a vocabulary of $K_{action}=$~5,656 discrete action tokens. For the spatial soft-labeling procedure, a neighbor weighting radius of $R=10$ cells and a Gaussian standard deviation of $\sigma=1.2$ are employed.
Furthermore, to enable hierarchical action generation for the coarse-to-fine (C2F) framework, two special tokens are introduced: the \textit{path goal token} and the \textit{waypoint goal token}.

\noindent\textbf{Training Details.}
We use the InternVL2-1B from the Mini-InternVL family~\cite{gao2024mini} as our main architecture.
The InternVL2-1B~\cite{gao2024mini} model consists of the
vision encoder InternViT-300M-448px (ViT)~\cite{chen2024internvl} and Qwen2-0.5B-Instruct~\cite{team2024qwen2} as the language model (LLM).
We train our model using the AdamW optimizer~\cite{kinga2015method} with a cosine learning rate schedule. The hyperparameters are set as follows: a base learning rate of 1e-4, weight decay of 0.1, $\beta_1=0.9$, $\beta_2=0.999$, and a dropout rate of 0.1. 
The model is trained for 30 epochs on 32 H20 GPUs with a batch size of 48. For model adaptation, we follow SimLingo~\cite{renz2025simlingo} and apply LoRA~\cite{hu2022lora} with a rank of 32 and $\alpha=64$.

During inference, we employ a Chain-of-Thought (CoT) approach. First, the model generates a textual rationale for the action. Conditioned on this commentary, it then predicts the final action sequence. This output consists of 20 geometric path tokens and 10 temporal waypoint tokens per frame.

\noindent\textbf{Unified understanding and generation.} We randomly concatenate a $(V, L, A)$ tuple to:~1) $[V, A, L]$ and supervise $L$ for action understanding, or 2) $[V, L, A]$ and supervise $A$ for action generation. Both are trained together with $\mathcal{L}_{total}$.

\begin{figure*}[!t]
    \centering
    \includegraphics[width=\textwidth]{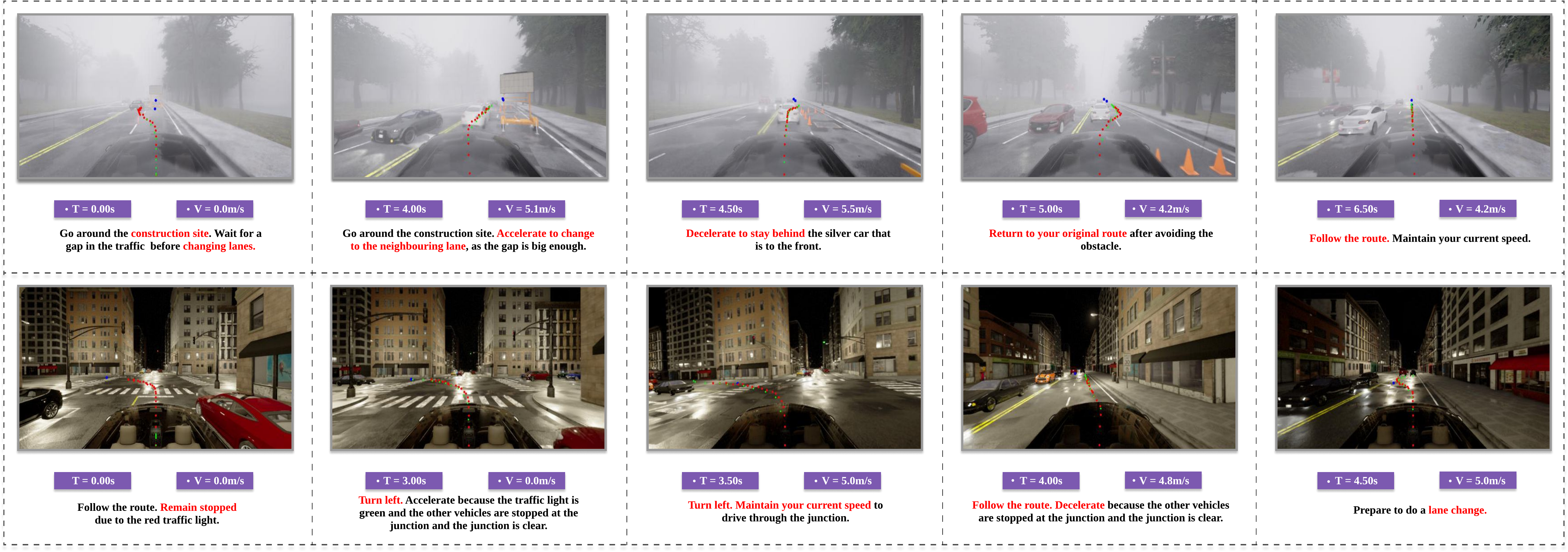}
    \vspace{-15pt}
    \caption{Visualization in challenging environment with various language instructions. The generated trajectory accurately adheres to the language instruction while remaining safe and feasible within the complex environment.}
    \label{fig:vis}
\end{figure*}
\vspace{-0.1cm}
\input{tables/tab_instruction}
\vspace{-0.1cm}
\input{tables/tab_language_driving_ablation}

\subsection{Results}
\vspace{-0.15cm}
\label{e-sec:results}
\noindent\textbf{Bench2Drive Results.}
Table~\ref{tab:main_combined} reports closed-loop metrics and multi-ability evaluations on Bench2Drive. 
LinkVLA archives the highest driving score and success rate while maintaining comparable efficiency and comfortness. 
Concretely, LinkVLA achieves driving score of 91.01 and an success rate of 74.55, surpassing the previous state of the art, SimLingo (85.07 DS, 67.27 SR), by 5.94 and 7.28 points, respectively, corresponding to relative gains of 6.98\% and 10.82\%. 
For efficiency, LinkVLA reaches 255.84, markedly exceeding earlier methods such as Orion (151.48) and AutoVLA (146.93). 
For comfortness, LinkVLA (34.62) modestly surpasses SimLingo (33.67) and substantially outperforms Orion (17.38).

The multi-ability results show robust driving behavior across diverse interaction scenarios. 
LinkVLA achieves the best scores in Merging (60.00), Overtake (80.00), Brake (93.33), and Traffic-Sign (83.68), and matches the top performance on Give-Way (50.00). 
The gains are especially pronounced in interaction-heavy and hazard-responsive skills, with improvements over SimLingo of 6.25 points in Merging, 11.11 in Overtake, and 11.66 in Brake. 
The overall multi-ability perfomance reaches 73.40, outperforming SimLingo (67.28) by 6.12 points (9.09\% relative) and Orion (54.72) by 18.68, indicating that LinkVLA is better for closed-loop driving performance.

\noindent\textbf{Inference Latency.}
Table~\ref{tab:latency} presents a comparative analysis of the latency-performance on the Bench2Drive benchmark.
Latency is quantified as the average inference time to generate one compact trajectory per frame. We omit the computational cost of the Chain-of-Thought (CoT) process, as its variable, query-dependent nature would introduce a confounding factor into the latency analysis.

Among the baselines, SimLingo is the fastest with a latency of 34 ms per step but achieves a Driving Score of 85.07.
In contrast, Orion (VAE) is slower at 65 ms.
Our auto-regressive (AR) variant achieves a high Driving Score of 90.66 but at a prohibitive latency of 361 ms.

In contrast, our proposed C2F method achieves a superior balance. It reduces latency from 361 ms to 48 ms, while simultaneously increasing the Driving Score to 91.01, the highest among all compared methods, achieving a 13.27-point higher Driving Score with 26\% lower latency than Orion. Compared to the fastest baseline, SimLingo, our method provides a 5.94-point performance gain with only a modest 14~ms increase in latency.

\subsection{Comparison Analysis}
\label{sec:comparison}

\noindent\textbf{Instruction-following Evaluation.} Our instruction-following evaluation, detailed in Table~\ref{tab:instruction-following}, shows that the progressive addition of action tokenization, coarse-to-fine (C2F) trajectory generation, and the action understanding objective for alignment~(short for \textit{align.}~in table) substantially improves performance. Compared to the baseline (ID 1, 70.11\% mean success), introducing tokenization alone (ID 2) boosts the overall success rate to 81.63\%. This gain is driven by near-perfect performance on the Stop task (99.88\%) and significant improvements in Lane Change (88.49\%) and Object centric (84.34\%). Subsequently adding C2F generation (ID 3) further enhances performance on tasks such as Faster (93.16\%), Target Speed (69.24\%), and Lane Change (95.45\%). Finally, incorporating the alignment module (ID 4) yields the most optimal and balanced performance. This final configuration achieves the highest mean success rate of 87.16\% and sets new peak scores for Faster (96.48\%), Target Speed (74.73\%), and Lane Change (97.42\%), demonstrating the surprising instruction-following ability of our method and the  effectiveness of our proposed action understanding objective.
\vspace{0.15cm}

\noindent\textbf{Language Ability Evaluation.}
Table~\ref{tab:language} indicates that across both benchmarks, enabling tokenization (ID 2) yields consistent gains over the baseline (ID 1) in \textit{SPICE}, \textit{BLEU}, and \textit{ROUGE-L}. 
Adding C2F (ID 3) further strengthens semantic fidelity: \textit{SPICE} rises on both DriveLM-VQA (to 71.3) and commentary (to 53.6).
Incorporating alignment training in addition to tokenization and C2F (ID 4) yields the best language understand performance in both VQA and commentary benchmarks, with all metrics improved.
Thanks to the unified token space design, our model has achieved more outstanding language ability.

\vspace{0.15cm}

\noindent\textbf{Ablation Experiment.}
Table~\ref{tab:driving_ablation} shows that tokenization yields the large improvement, increasing the driving score from 85.07\% to 89.57\% and the success rate from 67.27\% to 73.18\%. 
Adding C2F without alignment produces a marginal gain in driving score (to 89.85\%) and slightly reduces success rate (to 72.27\%). 
Incorporating alignment in addition to tokenization and C2F yields the best performance (driving score 91.01\%; success rate 74.55\%), surpassing all prior configurations.

\input{tables/tab_soft_label}
\noindent\textbf{Effect of Soft-labeling.}
Table~\ref{tab:soft_label} shows that adding soft-label supervision yields consistent gains in closed-loop performance. The driving score increases from 90.85 to 91.01 (by 0.16 points), and the success rate rises from 72.73\% to 74.55\% (by 1.82 points). 
These results suggest that soft-label supervision effectively leverages spatial prior and results in robust driving performance and task completion.

\input{tables/tab_tp_nc}
\noindent\textbf{Navigation Modalities.}
Table~\ref{tab:tp&nc} shows that the two navigation modalities deliver comparable closed-loop performance. It shows the ability of the LinkVLA to follow basic navigational commands and navigational GPS target points with the same model.

\noindent\textbf{Additional ablation Discussions.} 
Due to space limitation, we further provide more extensive ablation studies on action codebook size $K_{action}$, scaling factor $k$ in the log transformation, and the spread scale parameter $\sigma$ in the spatial soft-labeling in \textit{Supplementary Materials}.

%% file: tables/tab_main_ability.tex
\begin{table*}[!h]
\centering
\renewcommand{\arraystretch}{1.08}
\setlength{\tabcolsep}{2.9pt}
\small
\caption{\textbf{Main results and multi-ability in Bench2Drive}. * denote expert feature distillation.}
\label{tab:main_combined}
\begin{tabular}{@{}L{2.9cm} G{1.2cm} G{1.2cm} C{1.2cm} C{1.2cm} C{1cm} C{1cm} C{1cm} C{1cm} C{1cm} G{1cm}@{}}
\toprule
\multirow{2}{*}{Method} & \multicolumn{4}{c}{Closed-loop metrics $\uparrow$} & \multicolumn{6}{c}{Multi-Ability (\%) $\uparrow$} \\
\cmidrule(lr){2-5} \cmidrule(lr){6-11}
& DS & SR~(\%) & Efficiency & Comfort. & Merging & Overtake & Brake & Give-Way & Traffic-Sign & Mean \\
\midrule
TCP*~\cite{wu2022trajectory}   & 40.70 & 15.00 & 54.26 & 47.80 & 16.18 & 20.00 & 20.00 & 10.00 & 6.99 & 14.63 \\
TCP-ctrl*                      & 30.47 & 7.27  & 55.97 & \textbf{51.51} & 10.29 & 4.44  & 10.00 & 10.00 & 6.45 & 8.23 \\
TCP-traj*                      & 59.90 & 30.00 & 76.54 & 18.08 & 8.89  & 24.29 & 51.67 & 40.00 & 46.28 & 34.22 \\
ThinkTwice*~\cite{jia2023think} & 62.44 & 31.23 & 69.33 & 16.22 & 27.38 & 18.42 & 35.82 & 50.00 & 54.23 & 37.17 \\
DriveAdapter*~\cite{jia2023driveadapter} & 64.22 & 33.08 & 70.22 & 16.01 & 28.82 & 26.38 & 48.76 & 50.00 & 56.43 & 42.08 \\
\cmidrule[0.5pt]{1-11}
AD-MLP~\cite{zhai2023rethinking}       & 18.05 & 0.00  & 48.45 & 22.63 & 0.00  & 0.00  & 0.00  & 0.00  & 4.35  & 0.87 \\
UniAD-Tiny~\cite{hu2023planning} & 40.73 & 13.18 & 123.92 & 47.04 & 8.89  & 9.33  & 20.00 & 20.00 & 15.43 & 14.73 \\
UniAD-Base~\cite{hu2023planning}       & 45.81 & 16.36 & 129.21 & 43.58 & 14.10 & 17.78 & 21.67 & 10.00 & 14.21 & 15.55 \\
VAD~\cite{jiang2023vad}                & 42.35 & 15.00 & 157.94 & 46.01 & 8.11  & 24.44 & 18.64 & 20.00 & 19.15 & 18.07 \\
DriveTransformer~\cite{jia2025drivetransformer} & 63.46 & 35.01 & 100.64 & 20.78 & 17.57 & 35.00 & 48.36 & 40.00 & 52.10 & 38.60
 \\
Orion~\cite{fu2025orion}               & 77.74 & 54.62 & 151.48 & 17.38 & 25.00 & 71.11 & 78.33 & 30.00 & 69.15 & 54.72 \\
AutoVLA~\cite{zhou2025autovla}         & 78.84 & 57.73 & 146.93 & 39.33 & - & - & - & - & - & - \\
SimLingo~\cite{renz2025simlingo}       & 85.07 & 67.27 & \textbf{259.23} & 33.67 & 53.75 & 68.89 & 81.67 & 50.00 & 82.11 & 67.28 \\
\midrule
\rowcolor{lavender}
LinkVLA (Ours)                          & \textbf{91.01} & \textbf{74.55} & 255.84 & 34.62 & \textbf{60.00} & \textbf{80.00} & \textbf{93.33} & \textbf{50.00} & \textbf{83.68} & \textbf{73.40} \\
\bottomrule
\end{tabular}
\end{table*}

%% file: tables/tab_latency.tex
\begin{table}[h]
\centering
\renewcommand\arraystretch{1.2}
\setlength{\tabcolsep}{8pt}
\small
\caption{\textbf{Performance and Latency Comparison.} All metrics are evaluated on the CARLA benchmark. Latency is the average inference time per step, measured on H20 GPU.}
\label{tab:latency}

\begin{tabular}{@{}l l c c c@{}}
\toprule
ID & Method & Type & Latency$\downarrow$ & Driving Score$\uparrow$ \\
\midrule
1  & Orion~\cite{fu2025orion} & VAE & 65ms & 77.74 \\
2  & Simlingo~\cite{renz2025simlingo}  & MLP & 34ms & 85.07 \\
3  & Ours & AR & 361ms & 90.66 \\
\rowcolor{lavender}
4  & Ours & C2F & 48ms & 91.01 \\
\bottomrule
\end{tabular}
\end{table}

%% file: tables/tab_instruction.tex
\begin{table*}[!t]
\centering
\renewcommand\arraystretch{1.08}
\setlength{\tabcolsep}{5pt}
\small
\caption{\textbf{Instruction-following evaluation} on \textit{Action Dreaming} dataset~\cite{renz2025simlingo}. Align. refers to alignment with unified training.}
\label{tab:instruction-following}

\begin{tabularx}{0.9\textwidth}{@{}*{11}{>{\centering\arraybackslash}X}@{}}
\toprule
\multirow{3}{*}{ID} & \multirow{3}{*}{\makecell{Token}} & \multirow{3}{*}{C2F} & \multirow{3}{*}{\makecell{Align.}} & \multicolumn{7}{c}{Success Rate (\%) $\uparrow$} \\
\cmidrule(lr){5-11}
& & & & Faster & Slower & \makecell{Target \\ Speed} & \makecell{Lane \\ Change} & Object & Stop & Mean \\
\midrule
1    & \xmark & \xmark & \xmark & 81.42 & 61.83 & 66.27 & 75.53  & 74.69 & 60.93 & 70.11 \\
2    & \cmark & \xmark & \xmark & 88.44 & 65.24 & 63.37 & 88.49 & 84.34  & \textbf{99.88} & 81.63 \\
3    & \cmark & \cmark & \xmark & 93.16 & 55.86 & 69.24 & 95.45 & 85.38 & 92.14 & 81.87 \\
\rowcolor{lavender}
4    & \cmark & \cmark & \cmark & \textbf{96.48} & \textbf{65.57} & \textbf{74.73} & \textbf{97.42} & \textbf{91.41} & 97.34 & \textbf{87.16} \\
\bottomrule
\end{tabularx}
\end{table*}
\vspace{0.15cm}

%% file: tables/tab_language_driving_ablation.tex
\begin{table*}[!t]
\centering
\begin{minipage}[t]{0.48\textwidth}
\centering
\renewcommand\arraystretch{1.06}
\setlength{\tabcolsep}{2.8pt}
\small
\captionof{table}{\textbf{Language Ability} in DriveLM-VQA and commentary evaluation. S / B / R refers to \textit{SPICE} / \textit{BLEU} / \textit{ROUGE-L}}
\label{tab:language}
\begin{tabular}{@{}c c c c c c c c c c@{}}
\toprule
\multirow{2}{*}{ID} & \multirow{2}{*}{\makecell{Token}} & \multirow{2}{*}{C2F} & \multirow{2}{*}{\makecell{Align.}} & \multicolumn{3}{c}{DriveLM-VQA $\uparrow$} & \multicolumn{3}{c}{Commentary$\uparrow$} \\
\cmidrule(lr){5-7} \cmidrule(lr){8-10}
& & & & S & B & R & S & B & R \\
\midrule
1 & \xmark & \xmark & \xmark & 66.7 & 68.9 & 71.5 & 49.2 & 60.3 & 64.3 \\
2 & \cmark & \xmark & \xmark & 69.7 & 70.5 & 73.1 & 53.3 & 63.7 & 68.0 \\
3 & \cmark & \cmark & \xmark & 71.3 & 69.9 & 73.4 & 53.6 & 61.6 & 67.3 \\
\rowcolor{lavender}
4 & \cmark & \cmark & \cmark & \textbf{73.0} & \textbf{74.7} & \textbf{77.0} & \textbf{57.4} & \textbf{65.7} & \textbf{70.8} \\
\bottomrule
\end{tabular}
\end{minipage}
\hfill
\begin{minipage}[t]{0.48\textwidth}
\centering
\renewcommand\arraystretch{1.06}
\setlength{\tabcolsep}{2.5pt}
\small
\captionof{table}{\textbf{Closed-loop performance} ablation study with different components we proposed.}
\label{tab:driving_ablation}
\begin{tabular}{@{}c c c c c c@{}}
\toprule
\multirow{2}{*}{ID} & \multirow{2}{*}{\makecell{Token}} & \multirow{2}{*}{C2F} & \multirow{2}{*}{\makecell{Align.}} & \multicolumn{2}{c}{Closed-loop metrics $\uparrow$} \\
\cmidrule(lr){5-6}
& & & & Driving Score & Success Rate(\%) \\
\midrule
1 & \xmark & \xmark & \xmark & 85.07 & 67.27 \\
2 & \cmark & \xmark & \xmark & 89.57 & 73.18 \\
3 & \cmark & \cmark & \xmark & 89.85 & 72.27 \\
\rowcolor{lavender}
4 & \cmark & \cmark & \cmark & \textbf{91.01} & \textbf{74.55} \\
\bottomrule
\end{tabular}
\end{minipage}
\end{table*}

%% file: tables/tab_soft_label.tex
\begin{table}[!h]
\centering
\renewcommand\arraystretch{1}
\setlength{\tabcolsep}{8pt}
\small
\caption{Effect of soft label in close-loop evaluation.}
\label{tab:soft_label}

\begin{tabular}{@{}l c c@{}}
\toprule
Method & Driving Score$\uparrow$ & Success Rate(\%)$\uparrow$ \\
\midrule
Ours \textit{w/o.} soft label & 90.85 & 72.73 \\
Ours \textit{w/.} soft label  & 91.01 & 74.55  \\
\bottomrule
\end{tabular}
\end{table}

%% file: tables/tab_tp_nc.tex
\begin{table}[!h]
\centering
\renewcommand\arraystretch{1}
\setlength{\tabcolsep}{6pt}
\small
\caption{Effect of different navigation forms.}
\label{tab:tp&nc}

\begin{tabular}{@{}l c c@{}}
\toprule
Method & Driving Score$\uparrow$ & Success Rate(\%)$\uparrow$ \\
\midrule
GPS target point  & 91.01 & 74.55  \\
Navigation command  & 91.25 & 73.18 \\
\bottomrule
\end{tabular}
\end{table}

%% file: sec/5_conclusion.tex
\section{Conclusion}\label{sec:conclusion}
We present \modelbf, a novel VLA model that improves language-action alignment and efficiency. We achieve deep cross-modal consistency by unifying language and action tokens in a shared codebook and training a novel bidirectional captioning objective. This core strategy, coupled with a coarse-to-fine decoder that cuts latency by 86\%, delivers substantial gains in instruction following and driving performance on closed-loop benchmarks. Our work thus provides a practical path toward reliable and responsive language-guided agents for real-world deployment. 

%% file: sec/6_ref.tex
{
    \small
    \bibliographystyle{ieeenat_fullname}
    \bibliography{main}
}

%% file: sec/7_appendix.tex
\clearpage
\appendix
\renewcommand\thesection{\Alph{section}}
\renewcommand\thefigure{S\arabic{figure}}
\renewcommand\thetable{S\arabic{table}}
\renewcommand\theequation{S\arabic{equation}}
\setcounter{figure}{0}
\setcounter{table}{0}
\setcounter{equation}{0}
\setcounter{page}{1}
\maketitlesupplementary

\section*{Appendix}

\section{Action Tokenization}
\label{appendix:log}

\subsection{Log-Coordinate Transformation}
We visualize the log-transformed coordinate space to facilitate intuitive comparison. We devise a non-uniform quantization scheme that prioritizes precision near the ego-vehicle by first applying a non-linear transformation to the waypoint coordinates~$(x, y)$ along each axis independently.

\begin{figure}[h]
\centering
\includegraphics[width=0.85\linewidth]{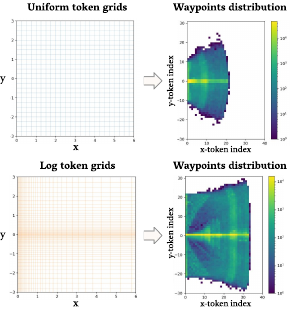}
\caption{Comparison of uniform and log token grids, with the corresponding waypoint distributions under each grid.}
\label{fig:f_1}
\end{figure}

% \newpage
\subsection{Number of Action Tokens}
We evaluate the effect of the number of action tokens on driving performance on the Bench2Drive~\cite{jia2024bench} benchmark. To this end, we adopt a non-uniform quantization scheme that prioritizes precision near the ego-vehicle by first applying a non-linear transformation to the waypoint coordinates $(x, y)$ along each axis independently. Specifically, each coordinate $z \in \{x, y\}$ is transformed using a signed logarithmic function:
\begin{equation}
z' = \operatorname{sign}(z) \cdot \log\!\big(1 + k \cdot |z|\big).
\label{eq:loggrid}
\end{equation}

We then vary the symmetric logarithmic (symlog) scaling factor $k$ from 5.0 to 10.0, which increases the number of action tokens from 5{,}656 to 7{,}245. The parameter $k$ controls the mapping from physical coordinates (both $x$ and $y$, in meters) to the transformed space, determining the degree of compression prior to binning and, in turn, the effective bin widths in the original coordinate space.

\begin{table}[!h]
\centering
\renewcommand\arraystretch{1}
\setlength{\tabcolsep}{8pt}
\small
\caption{Effect of the number of action tokens (controlled by $k$) in closed-loop evaluation~\cite{dosovitskiy2017carla}.}
\label{tab:num_tokens}
\begin{tabular}{@{}l c c@{}}
\toprule
Method & Driving Score$\uparrow$ & Success Rate (\%)$\uparrow$ \\
\midrule
k = 5.0  (5{,}656 tokens) & 91.01 & 74.55 \\
k = 10.0~(7{,}245 tokens) & 89.85 & 70.45 \\
\bottomrule
\end{tabular}
\end{table}

\subsection{Spatial Soft-Labeling}
We evaluate the spread scale in the spatial soft-labeling on driving performance on the Bench2Drive~\cite{jia2024bench} benchmark by varying the spread parameter $\sigma$ from 1.2 to 3.0.
Larger $\sigma$ yields softer targets by broadening the Gaussian smoothing and distributing probability over a wider neighborhood.
Specifically, for a ground-truth token $a_{gt}$, we construct the target distribution $q(a)$ over all action tokens $a \in \mathcal{C}_{\text{action}}$ as a normalized 2D Gaussian centered at the coordinates of $a_{gt}$:
\begin{equation}
    q(a) = \frac{1}{Z} \exp\left(-\frac{\| \text{pos}(a) - \text{pos}(a_{gt}) \|_2^2}{2\sigma^2}\right),
    \label{eq:soft_target}
\end{equation}
where $\text{pos}(a)$ maps an action token to its 2D coordinates in the spatial grid, $\| \cdot \|_2^2$ denotes the squared Euclidean distance, $\sigma$ is a hyperparameter controlling the spread of the distribution, and $Z$ is a normalization constant ensuring $\sum_{a \in \mathcal{C}_{\text{action}}} q(a) = 1$.

\begin{table}[!h]
\centering
\renewcommand\arraystretch{1}
\setlength{\tabcolsep}{8pt}
\small
\caption{Effect of the spread scale parameter $\sigma$ in the spatial soft-labeling in closed-loop evaluation~\cite{dosovitskiy2017carla}.}
\label{tab:soft_label}

\begin{tabular}{@{}l c c@{}}
\toprule
Method & Driving Score$\uparrow$ & Success Rate (\%)$\uparrow$ \\
\midrule
$\sigma = 1.2$ & 91.01 & 74.55 \\
$\sigma = 3.0$ & 89.73 & 69.55 \\
\bottomrule
\end{tabular}
\end{table}

\newpage

\section{Dataset}
\label{appendix:dataset}
\subsection{Action Dreaming}
We conduct experiments on the \textit{Action Dreaming}~\cite{renz2025simlingo} dataset and its offline, nonreactive simulator, which generates alternative ego-vehicle trajectories and assesses their feasibility with respect to collision avoidance and traffic-rule compliance.
Ego-trajectory prediction is implemented with a kinematic bicycle model controlled by PID controllers (PDM-lite~\cite{sima2024drivelm}), driven either by perturbed ground-truth actions or by PID commands computed from predefined path waypoints and target speeds.
The dataset provides simulator states and short-horizon forecasts for dynamic objects to enable collision checking. The simulator supports several modes (Objects/Collision, Faster, Slower, Target Speed, Lane Changes, Stop) to induce diverse behaviors and trajectories.
we use Success Rate as the metric. 
Each category has its own definition of success, which we detail in the following:

\begin{itemize}[leftmargin=*, itemsep=0.25em]
\item Objects (Collision): This describes the task of driving towards or crashing into specific objects. 
The path is evaluated first. 
If the path of the expert trajectory and the ground truth dreamer trajectory is different (Average Displacement Error $ADE > 1.0$) it is counted as success if the predicted path is closer to the ground truth dreamer path than to the expert path ($ADE_{pred2expert} > ADE_{pred2dreamer}$). If the dreamer path is nearly identical to the expert path ($ADE < 1.0$) the instruction is about correct speed predictions (\textit{e.g.}, if the instruction is “drive towards a dynamic object” it is important to get the speed right and not just the path). The success is then defined as $ADE_{pred2dreamer} < 1.0$ and the average predicted speed is within 30\% of the ground truth dreamer speed.

\item Faster (Speed up): For each predicted speed waypoint, derive target speeds for future timesteps and fit a linear regression to obtain the slope \( s \). Let \( v \) denote the current ego speed at the start of the sequence. Success is defined as \( s > 0.05\,v \).

\item Slower (Slow down): Same computation as Faster. Success is defined as \( s < -0.05\,v \).

\item Target Speed: Since the target speed may not be reached in the prediction horizon of the waypoints, predictions are compared with the ground truth actions instead of directly comparing to the target speed. Two rules define success: First, if the predicted target speed inferred from the last two waypoints is in a 20\% range of the instructed target speed. Second, if the predicted target speed inferred from the last two waypoints is in the 20\% range of the speed of the last two waypoints of the ground truth speed waypoints. This can be different from the instructed target speed due to limitations in the acceleration rates of the vehicle.

\item Lane Changes: Compare the final waypoint of the predicted path with the final waypoints of the ground-truth dreamer path and the ground-truth expert path. Success is defined when the predicted final location is closer to the dreamer’s final location than to the expert’s final location.

\item Stop: Success is defined when the minimum predicted speed over the sequence is below \( 0.1\,\mathrm{m/s} \).
\end{itemize}

\subsection{VQA-DriveLM}
The VQA data from SimLingo~\cite{renz2025simlingo} are sourced from the DriveLM-Carla~\cite{sima2023drivelm} dataset and generated using its data-creation pipeline. Question–answer pairs are extracted from the adopted dataset rather than the original DriveLM release; the training split contains 28M QA pairs over 1M frames in Town 12. Evaluation follows DriveLM keyframe selection to focus on informative frames, and the validation split is balanced across answer types. Labels are heuristically auto-generated, and the dataset includes GPT-4–based paraphrase augmentation (up to 20 variants per QA) to mitigate phrase-level overfitting, with variants sampled at load time.

\subsection{Commentary}
The commentary labels in SimLingo~\cite{renz2025simlingo} are automatically generated from a subset of saved simulator state using heuristic, template-based rules. Each label comprises: (1) a route action with justification—default “Follow the route,” replaced only for scenarios requiring lane deviation (\textit{e.g.}, obstacle, encroaching vehicle), with phase-specific templates for before/during/after the deviation; (2) a speed action categorized as remain stopped, stop now, maintain (or maintain reduced) speed, increase speed, or slow down, with a special “Wait for a gap before changing lanes” case when stationary prior to a deviation; and (3) a speed reason derived from IDM~\cite{treiber2000congested} features that identify the leading object (vehicle/pedestrian/static control) and its attributes/state, from which concise rationales are composed (\textit{e.g.}, due to pedestrian crossing, behind a red SUV, because the light is red/green). When near a junction, an additional notice summarizes other vehicles’ positions and motions (\textit{e.g.}, junction clear, vehicle moving away, oncoming traffic).

\section{Qualitative Results}
\label{appendix::qualitative results}
\begin{figure*}[t]
\centering
\includegraphics[height=0.95\textheight, width=0.95\linewidth, keepaspectratio]{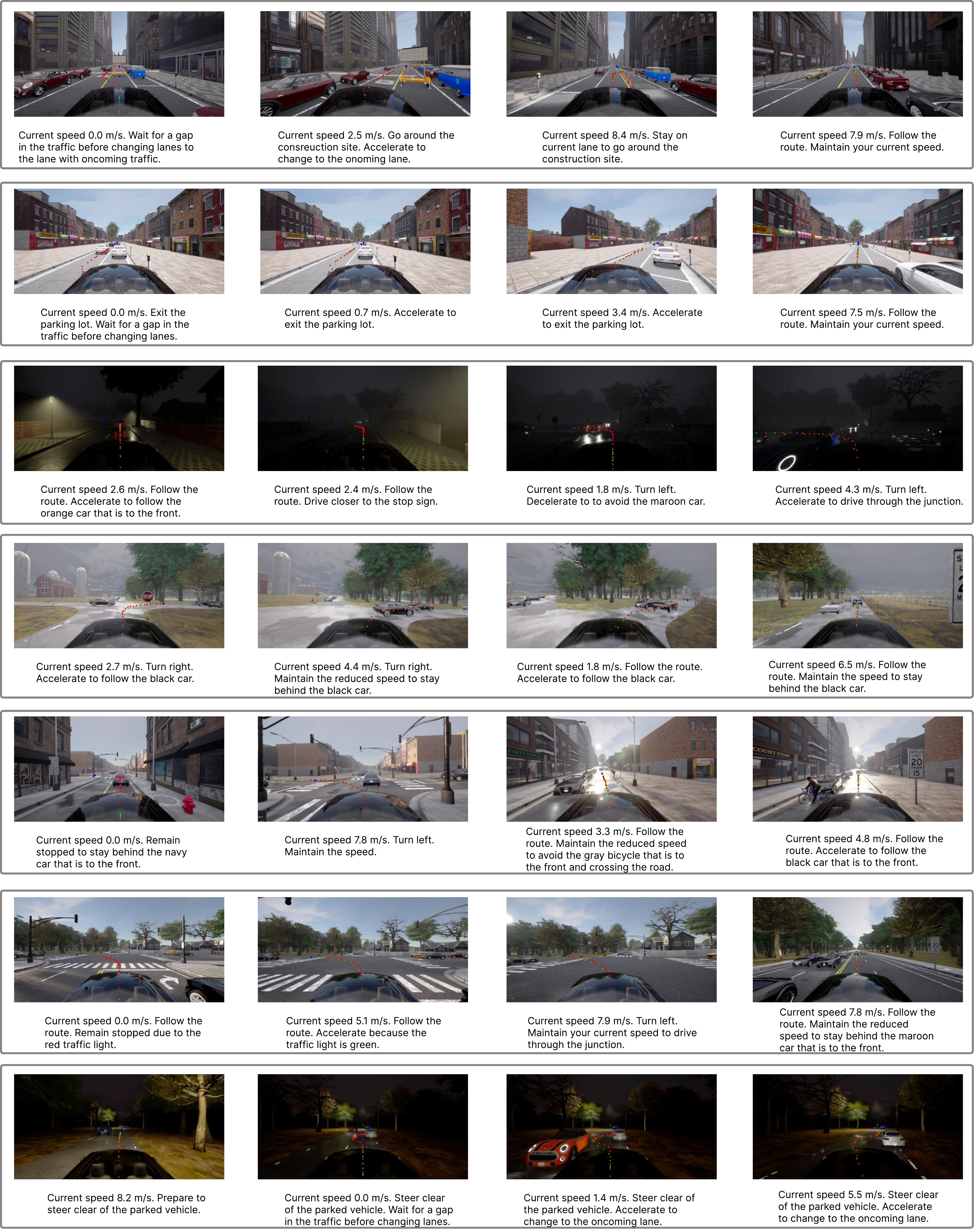}
\caption{Qualitative results of our proposed model during closed-loop evaluation in the CARLA simulator. The figure showcases representative driving scenarios, such as navigating intersections and avoiding obstacles.}
\label{fig:f_vis}
\end{figure*}

Figure~\ref{fig:f_vis} provides further qualitative results from the closed-loop evaluation, illustrating our model's performance in a variety of driving scenarios.

\newpage